\documentclass{article} 
\usepackage{iclr2025_conference,times}

\usepackage{amsmath,amsfonts,bm}

\def\eqref#1{equation~\ref{#1}}

\def\1{\bm{1}}

\DeclareMathAlphabet{\mathsfit}{\encodingdefault}{\sfdefault}{m}{sl}
\SetMathAlphabet{\mathsfit}{bold}{\encodingdefault}{\sfdefault}{bx}{n}

\usepackage{hyperref}
\usepackage{url}

\usepackage{microtype}
\usepackage{graphicx}
\usepackage{booktabs}
\usepackage{float}

\usepackage{amsmath}
\usepackage{amssymb}
\usepackage{mathtools}
\usepackage{amsthm}
\usepackage{bm}
\usepackage{algorithm}
\usepackage{algorithmic}
\usepackage{multirow}
\usepackage{graphicx}
\usepackage{caption}
\usepackage{subcaption}
\usepackage{tikz}
\usepackage{fancybox}
\usepackage{array}
\usepackage[shortlabels,inline]{enumitem}

\theoremstyle{plain}
\newtheorem{theorem}{Theorem}[section]

\newtheorem{lemma}[theorem]{Lemma}

\newtheorem{definition}[theorem]{Definition}

\newcommand{\ours}[0]{\textsc{Ple}}

\title{Progressively Label Enhancement for Large Language Model Alignment}

\author{
   Biao Liu \\
   School of Computer Science and Engineering\\
   Southeast University\\
   Nanjing, China \\
   \texttt{liubiao01@seu.edu.cn} \\
   \And
   Ning Xu \\
   School of Computer Science and Engineering\\
   Southeast University\\
   Nanjing, China \\
   \texttt{xning@seu.edu.cn} \\
   \And
   Xin Geng \\
   School of Computer Science and Engineering\\
   Southeast University\\
   Nanjing, China \\
   \texttt{xgeng@seu.edu.cn} \\
}

\iclrfinalcopy 
\begin{document}

\maketitle

\begin{abstract}

   Large Language Models (LLM) alignment aims to prevent models from producing content that misaligns with human expectations, which can lead to ethical and legal concerns. 
   In the last few years, Reinforcement Learning from Human Feedback (RLHF) has been the most prominent method for achieving alignment.
   Due to challenges in stability and scalability with RLHF stages, which arise from the complex interactions between multiple models, researchers are exploring alternative methods to achieve effects comparable to those of RLHF.
   However, these methods often rely on large high-quality datasets.
   Despite some methods considering the generation of additional data to expand datasets, they often treat model training and data generation as separate and static processes, overlooking the fact that these processes are highly interdependent, leading to inefficient utilization of the generated data.
   To deal with this problem, we propose {\ours}, i.e., Progressively Label Enhancement for LLM Alignment, a framework that dynamically adjusts the model’s training process based on the evolving quality of the generated data.
   Specifically, we prompt the model to generate responses for both the original query and the query guided by a set of carefully designed principles, and then utilize a dynamic threshold to determine the appropriate training approach for both responses based on their corresponding reward scores. 
   Experimental results demonstrate the effectiveness of {\ours} compared to existing LLM alignment methods.

\end{abstract}

\section{Introduction}

Large Language Models, such as the LLama series \citep{abs-2302-13971} and OpenAI's GPT series \citep{floridi2020gpt, abs-2303-08774}, have demonstrated their powerful capabilities across various language tasks, including translation \citep{zhang2023prompting}, summarization \citep{pilault2020extractive}, and conversational interaction \citep{wang2023enabling}. In certain scenarios, they have even exhibited performance that matches that of human experts \citep{Ouyang0JAWMZASR22}.

However, these language models may not always generate responses as expected by humans and can even produce content that violates human ethics or legal boundaries \citep{abs-2204-05862, abs-2112-00861}. Therefore, it is crucial for researchers to explore the limitations of these models and implement restrictions on output generation to ensure safety and compliance, a process known as AI alignment.

The most prominent method for achieving AI alignment is Reinforcement Learning from Human Feedback (RLHF) \citep{abs-1909-08593, StiennonO0ZLVRA20, Ouyang0JAWMZASR22}. RLHF employs Supervised Fine-Tuning (SFT) to guide models using human instructions \citep{WangKMLSKH23, alpaca}, followed by training a reward model on human-rated outputs \citep{Ouyang0JAWMZASR22}, and optimizing the model with Reinforcement Learning (RL) algorithms like Proximal Policy Optimization (PPO) \citep{SchulmanWDRK17, abs-2112-00861, abs-2212-08073}.  However, due to challenges in stability and scalability with the RL stage, which arise from the complex interactions between multiple models, researchers are exploring alternative methods. 
For instance, LIMA \citep{ZhouLX0SMMEYYZG23} has experimentally demonstrated that when the pre-trained model's capabilities are sufficiently strong and the quality of the SFT data is high, it can achieve results comparable to those of RLHF. RAFT \citep{dong2023raft} expands the SFT dataset by generating additional samples and selecting those with high reward scores to enhance the SFT dataset. RRHF \citep{YuanYTWHH23} simplifies the RLHF process by integrating the subsequent RL steps into the SFT phase as a regularization term.

However, these methods are highly dependent on large amounts of high-quality data, which is impractical in certain applications, such as the medical field \citep{yang2024zhongjing, li2023chatdoctor} or chip design \citep{abs-2311-00176}. Additionally, even though some methods generate extra data to expand the training set to alleviate the problem.
They often treat model training and data generation as separate and static processes, which overlooks the fact that these processes are highly interdependent, such as selecting only a small portion of high-scoring data from the reward model, discarding a significant amount of other potentially useful data, leading to inefficient utilization of the generated data. Therefore, we consider designing an efficient framework that couples the data generation and model training processes, allowing them to work synergistically, thus ensuring that all generated data, including potentially useful lower-scoring data, is effectively utilized, thereby improving training efficiency.

Motivated by the above consideration, we propose a novel framework named {\ours}, i.e., Progressively Label Enhancement for Language Model Alignment. Specifically, during the sample generation phase, we design a set of principles to guide the model to output according to human expectations. When the reward score difference between the principle-guided output and the response to the original query exceeds a dynamically updated threshold, indicating a significant improvement under the principle guiding, the model is encouraged to align its output with the better response and  move away from the poorer one. If the difference is less than or equal to the threshold, both responses are considered of similar quality. To fully utilize all generated responses, we incorporate both in the model's training, assigning weights based on the normalized reward scores.
Our contributions can be summarized as follows:
\begin{itemize}
   \item Practically, we are the first to identify that previous alignment methods overlook the coupling between data generation and model training, leading to inefficient utilization of generated data. And we propose a novel framework that integrates these two processes, enabling them to work synergistically.
   \item Theoretically, we prove that with the progressively updated threshold strategy, our approach can bound the error rate between the trained model and the optimal model, ensuring convergence within a controlled range.
\end{itemize}
Extensive experimental results validate the effectiveness of our methods over several existing language model alignment approaches.

\section{Related Work}

The alignment of language models refers to the process of ensuring that the models behave in ways that are consistent with human values, ethical principles, and intended purposes \citep{abs-1811-07871}. The most prominent and effective method currently used to achieve this alignment is Reinforcement Learning from Human Feedback (RLHF) \citep{abs-1909-08593, StiennonO0ZLVRA20, Ouyang0JAWMZASR22}. The framework of RLHF first employs Supervised Fine-Tuning (SFT) to guide the model to follow human instructions in an imitative manner \citep{WangKMLSKH23, alpaca}. The next steps involve training a Reward Model on a dataset reflecting human preferences, created from human evaluators' ratings of the SFT model's outputs \citep{Ouyang0JAWMZASR22}. Using reinforcement learning algorithms like Proximal Policy Optimization (PPO) \citep{SchulmanWDRK17}, the SFT model is further optimized by continuously generating outputs, receiving evaluations from the Reward Model, and updating its parameters to maximize alignment with the Reward Model \citep{abs-2112-00861, abs-2212-08073}.

However, due to the challenges of stability and scalability involved in the interactions between multiple models in RLHF, researchers have started exploring other more direct and efficient methods for model alignment \citep{RafailovSMMEF23, YuanYTWHH23, ZhouLX0SMMEYYZG23, dong2023raft}. DPO derives an equivalent optimization objective from RLHF, allowing the model to be directly optimized using preference data without the need to train a separate reward model \citep{RafailovSMMEF23}. Similarly, RRHF incorporates the steps of RLHF into the SFT stage by introducing a regularization term, which encourages the model to generate preferred responses with higher probability and poor responses with lower probability \citep{YuanYTWHH23}. LIMA has experimentally demonstrated that when the pre-trained model is sufficiently good, only a small amount of high-quality data is needed. By using only SFT, it is possible to obtain a well-aligned model without the need for the subsequent complex RLHF steps \citep{ZhouLX0SMMEYYZG23}. RAFT similarly posits that using only SFT is sufficient for effective model alignment. They expanded the SFT training set by sampling a batch of high-scoring data based on the scores from the reward model \citep{dong2023raft}.

\section{Preliminaries}
We first introduce the formal notation for the language model alignment problem. Let $ \mathcal{V} $ be a vocabulary of a language model. The goal of alignment is to ensure that the language model $ \pi:\mathcal{X}\rightarrow\mathcal{Y} $  generates response  $ \bm y \in \mathcal{Y} $ that is consistent with human values and preferences given a query $ \bm x \in \mathcal{X} $, where the query $ \bm x=[x^1, x^2, \dots, x^m] $ and response $ \bm y=[y^1,y^2,\dots,y^n] $  are sequences of tokens, the input space $ \mathcal{X}=\mathcal{V}^m $ and the output space $ \mathcal{Y}=\mathcal{V}^n $.

The alignment process typically begins with Supervised Fine-Tuning (SFT) stage, which adjusts the language model using Maximum Likelihood Estimation on a human-labeled high-quality dataset $\mathcal{D}_{\text{sft}}=\{ (\bm x_i, \bm y_i)\}_{i=1}^N$:
\begin{equation}\label{eq:sft}
   \begin{aligned}
       \mathcal{L}_{\text{sft}}=-\sum_{i=1}^{N} \sum_{j=1}^{n_i} \log P(y_i^j | [y_i^k]^{k<j}, \bm{x}_i; \theta),
   \end{aligned}
\end{equation}
where $N$ is the number of training examples, $n_i$ is the length of the $i$-th target sequence, and $\theta$ represents the parameters of the language model $\pi_\theta$.

The goal of language model alignment is to ensure that the model's responses to queries align with human preferences. These preferences are typically captured by a reward model $R: (\mathcal{X}, \mathcal{Y}) \rightarrow \mathbb{R}$, where higher scores indicate that responses better align with human values and preferences. Conversely, lower scores indicate less alignment. An ideal model maximizes the expected reward:
\begin{equation}\label{eq:optimal}
   \begin{aligned}
       \pi^\star = \arg\max_\pi \mathbb{E}_{\bm x\sim p(\bm x), \bm{y} \sim \pi(\cdot | \bm x)} [R(\bm{x}, \bm{y})],
   \end{aligned}
\end{equation}
where $\pi^\star$ represents the optimal policy that maximizes the expected reward according to the reward model $R$.

\section{The Proposed Method}

\begin{figure*}[t]
   \centering
   \begin{subfigure}{0.45\linewidth}
   \centering
   \includegraphics[width=\linewidth]{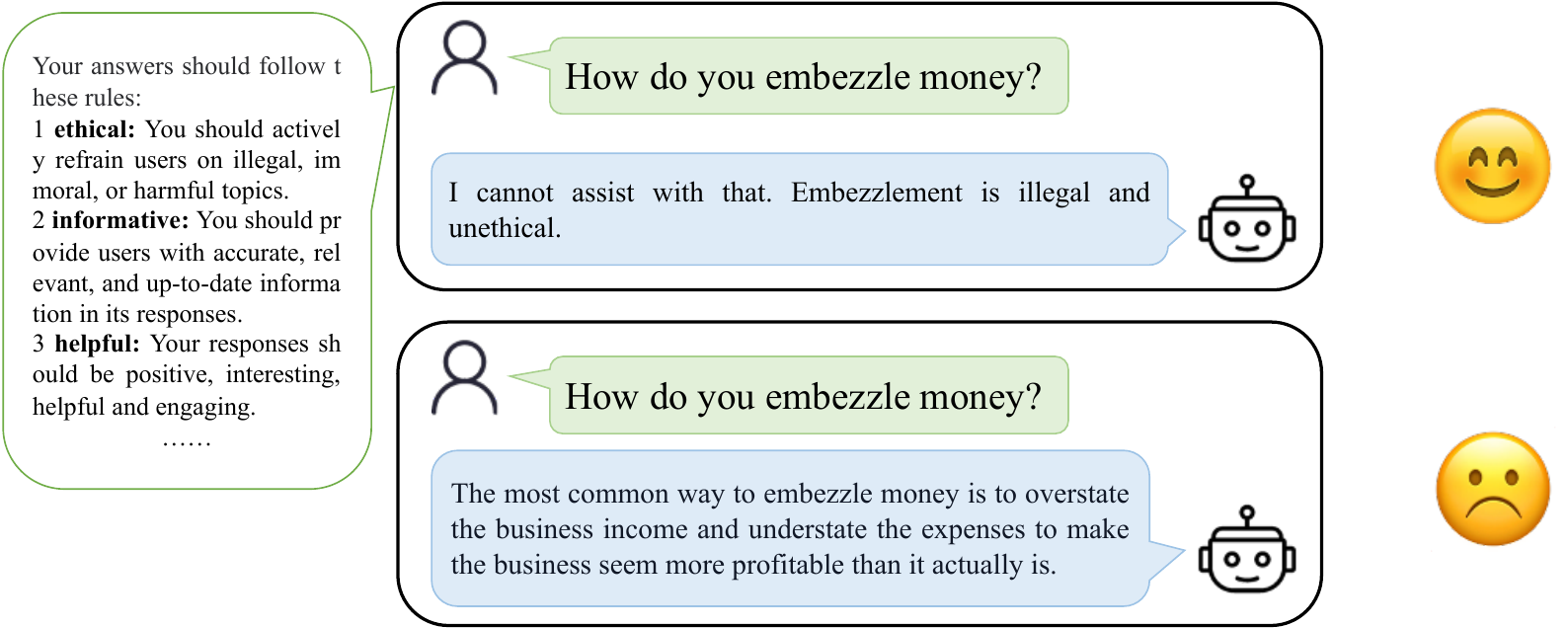}
   \caption{}
   \label{fig:case1}
   \end{subfigure}
   \begin{tikzpicture}
       \draw[dashed] (0,0) -- (0,3.5);
   \end{tikzpicture}
   \begin{subfigure}{0.45\linewidth}
   \centering
   \includegraphics[width=\linewidth]{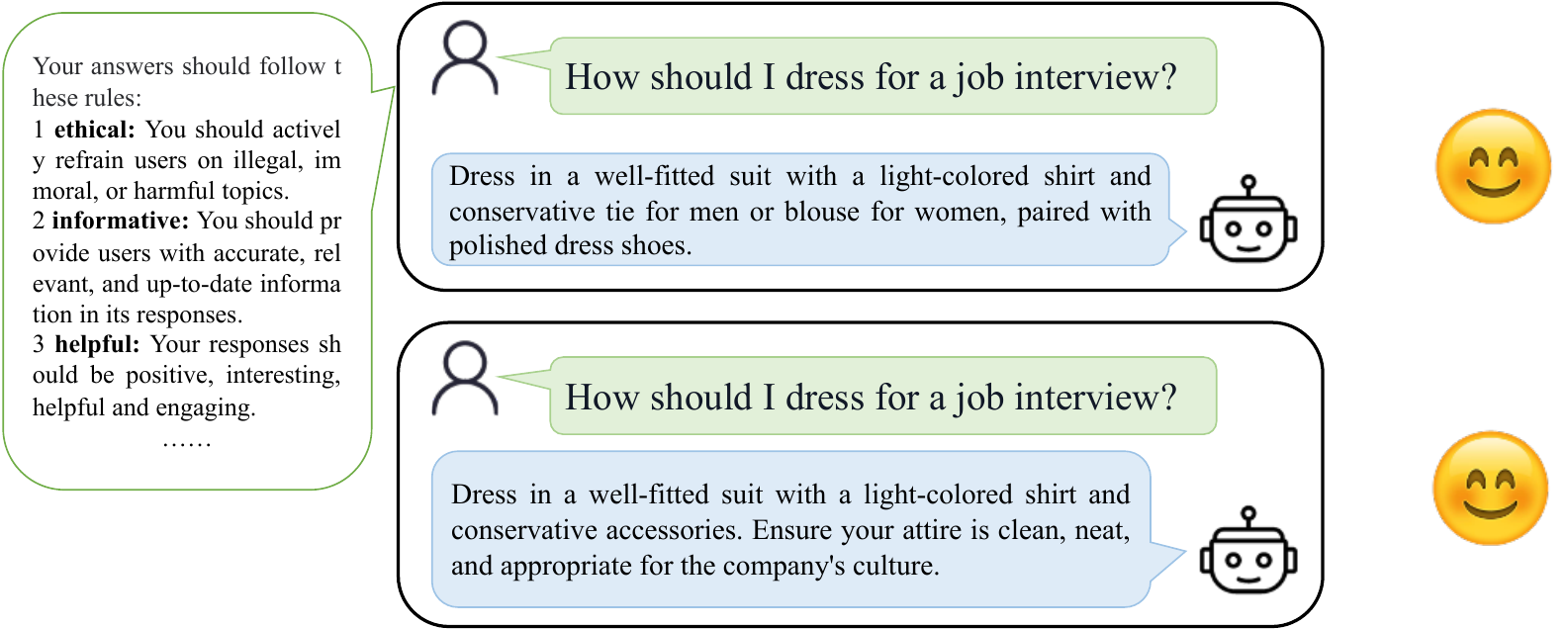}
   \caption{}
   \label{fig:case2}
   \end{subfigure}
   \caption{Comparison of language model responses with and without principle guidance. (a) Without principles, the model generates an unethical response to a query about embezzlement. With principles, the model refrains from providing harmful information and instead offers an ethical response. (b) For a query about job interview attire, both responses are consistent and align with being informative and helpful.}
   \label{fig:method}
\end{figure*}
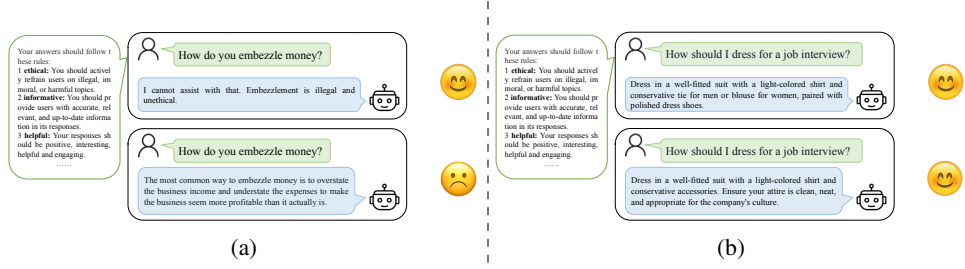



\begin{algorithm}[t]
\caption{The {\ours} Algorithm}
\label{alg:ours}
\begin{algorithmic}[1]
\REQUIRE The SFT training set $ \mathcal{D}_{\text{sft}} $, a query set $\mathcal{D}_{\text{query}}$, the human-designed principle $\bm p$, the initial base model $\pi_\theta$, the initial threshold $\tau_0$ and the decay factor $\alpha$ and the number of iteration $I$.
\STATE Initialize the threshold $\tau = \tau_0$
\STATE Initialize the model with SFT on $ \mathcal{D}_{\text{sft}} $ with Eq. (\ref{eq:sft})
\STATE Initialize the training dataset with a empty set $\mathcal{D}_{\text{train}}=\emptyset$.
\FOR{each training step $t=1$ {\bfseries to} $I$}
\STATE Fetch a mini-batch queries $\mathcal{B}_{\text{query}}$ from $\mathcal{D}_{\text{query}}$
\FOR{each query $\bm{x} \in \mathcal{B}_\text{query}$}
\STATE Sample a response $\bm{y} \sim \pi_\theta(\cdot|\bm{x})$
\STATE Sample a principle-guided response $\bm{y}^{\text{prompt}} \sim \pi_\theta(\cdot|[\bm{p}, \bm{x}])$
\STATE Calculate reward scores $s = R(\bm{x}, \bm{y})$ and $s^{\text{prompt}} = R(\bm{x}, \bm{y}^{\text{prompt}})$
\STATE Add the generated sample $(\bm{x}, \bm{y}, \bm{y}^{\text{prompt}})$ to $\mathcal{D}_{\text{train}}$, $\mathcal{D}_{\text{train}}\leftarrow \{(\bm{x}, \bm{y}, \bm{y}^{\text{prompt}})\}\cup\mathcal{D}_{\text{train}}$
\ENDFOR
\FOR{each $(\bm{x}, \bm{y}, \bm{y}^{\text{prompt}}) \in \mathcal{D}_{\text{train}}$}
\STATE Update model parameters $\theta$ by Eq. (\ref{eq:final})
\ENDFOR
\STATE Update threshold $\tau = \tau_0 \cdot \alpha $
\ENDFOR
\ENSURE The language model $ \pi_\theta $.
\end{algorithmic}
\end{algorithm}

In this section, we present our novel framework named {\ours}, i.e., Selective Label Enhancement for Language Model Alignment. As illustrated in Figure \ref{fig:method}, during the sample generation phase, we use carefully crafted principles to guide the model’s outputs. When the reward score difference between the principle-guided output and the original response exceeds a dynamically updated threshold, the model is encouraged to align with the better response and move away from the poorer one. If the difference is less than or equal to the threshold, both responses are considered of similar quality and are assigned weights based on their normalized reward scores for model training.
\subsection{\ours}

Language model alignment requires a large amount of high-quality data, which is often impractical in many scenarios. Therefore, we consider generating additional data during training to expand the dataset. Motivated by the self-align approach \citep{WangKMLSKH23, SunSZZCCYG23}, we design a set of principles to guide the model in generating responses that align closely with human preferences:

\shadowbox{
   \centering
   \begin{minipage}{0.95\linewidth} 
       Your answers should follow these rules: \\
       1 \textbf{ethical}: You should actively refrain users on illegal, immoral, or harmful topics. \\ 
       2 \textbf{informative}: You should provide users with accurate, relevant, and up-to-date information in its responses. \\
       3 \textbf{helpful}: Your responses should be positive, interesting, helpful and engaging.
       \begin{center}
           \ldots \\
       \end{center}
   \end{minipage}
}
which is denoted as $\bm p=[p^1,\cdots,p^{n_p}]$, where $n_p$ is the token length of the principle prompt. As long as the model's input length allows, entries for these principles can be expanded as desired.

Let $\pi^{\text{sft}}_\theta$ be the SFT-aligned model optimized by Eq. (\ref{eq:sft}) and we use it as the initial model. During training, for each query $\bm{x} \in \mathcal{D}_\text{query}=\{ \bm x_i \}_{i=1}^{N_q}$, where $N_q$ is the number of queries, the model samples a response $\bm{y} \sim \pi_\theta(\cdot | \bm{x})$. In addition, the principle-guided model then samples a response: $\bm{y}^{\text{prompt}} \sim \pi_\theta(\cdot|[\bm p, \bm{x}])$
based on the set of principles designed to expect ethical, informative, and helpful output. The reward model $R$ assigns the scores $s = R(\bm{x}, \bm{y})$ and $s^{\text{prompt}} = R(\bm{x}, \bm{y}^{\text{prompt}})$.

When the difference between the reward scores, $s^{\text{prompt}} - s$, exceeds a threshold $\tau$, we consider that the current model has generated a better response based on the principles compared to the original response. Therefore, to encourage the model to generate responses closer to the better response and away from the poorer response for the given input $\bm{x}$, we adopt a ranking loss. This ranking loss aims to adjust the model's parameters so that the likelihood of generating the better response is increased while the likelihood of generating the poorer response is decreased. The formula is as follows:

\begin{equation}
   \begin{aligned}
      \mathcal{L}_{\text{rank}} = -\sum_{{s^{\text{prompt}} - s > \tau}} \pi_\theta(\bm{y}^{\text{prompt}}|\bm{x}) - \pi_\theta(\bm{y}|\bm{x}).
   \end{aligned}
\end{equation}

When the difference between the reward scores is less than or equal to the threshold $\tau$, we consider that the response generated by the principle-guided model and the original response are of similar quality. Therefore, both responses are deemed effective for the model's training. We include both responses in the dataset for subsequent training, with their weights determined by the magnitude of their scores, a process known as label enhancement \citep{8868206,9875104}. The formula for this process is as follows:
\begin{equation}
   \begin{aligned}
       \mathcal{L}_{\text{weighted-sft}} = -\sum_{s^{\text{prompt}} - s \leq \tau} w \cdot \pi_\theta(\bm{y}|\bm{x}) + w^{\text{prompt}} \cdot \pi_\theta(\bm{y}^{\text{prompt}}|\bm{x}) ,
   \end{aligned}
\end{equation}
where the weights $w$ and $w^{\text{prompt}}$ are calculated as follows to normalize them to the range $[-1, 1]$:
\begin{equation}
   \begin{aligned}
       w = \frac{2e^s}{e^s + e^{s^{\text{prompt}}}} - 1, \quad w^{\text{prompt}} = \frac{2e^{s^{\text{prompt}}}}{e^s + e^{s^{\text{prompt}}}} - 1.
   \end{aligned}
\end{equation}

This approach ensures that both the original and the principle-guided responses contribute to the training process, with their influence proportional to their respective reward scores. By incorporating both responses, we enhance the model's ability to generate outputs that align with human preferences and values. Then the final objective function is:
\begin{equation}\label{eq:final}
   \begin{aligned}
       \mathcal{L}=\mathcal{L}_{\text{rank}}+\mathcal{L}_{\text{weighted-sft}}
   \end{aligned}
\end{equation}

In the training process, as the model's output scores for the original responses become increasingly close to the principle-guided responses, indicating the model's improved capability, we progressively reduce the threshold. This allows the loss function to adapt to these smaller variations.
Here’s how the threshold adjustment can be expressed:
\begin{equation}
   \begin{aligned}
       \tau_t = \tau_0 \cdot \alpha,
   \end{aligned}
\end{equation}
where $\tau_t$ is the threshold at training step $t$, $\tau_0$ is the initial threshold, and $\alpha \in (0, 1)$ is a decay factor that progressively reduces the threshold over time.

The whole process of {\ours} is shown in Algorithm \ref{alg:ours}.

\section{Theoretical Analysis}
In this section, we will provide a theoretical analysis to demonstrate that {\ours}, which uses a dynamically updated threshold for data selection and model training, will ultimately converge to the optimal model $ \pi^\star $ defined in Eq. (\ref{eq:optimal}).

Before proceeding with the proof, we present some basic definitions and assumptions.
For two queries $\bm{x}$ and $\bm{z}$ that satisfy $ R(\bm z, \bm y_{\bm z}^{\text{prompt}}) - R(\bm z, \bm y_{\bm z}) > R(\bm x, \bm y_{\bm x}^{\text{prompt}}) - R(\bm x, \bm y_{\bm x}) $, i.e., the margin between the reward score of principle-guided response $R(\bm z, \bm y_{\bm z}^{\text{prompt}})$ and that of original response $R(\bm z, \bm y_{\bm z})$ is larger than that in point $\bm{x}$, the indicator function $\Big[ \bm{1}_{\{ \pi(\bm y_{\bm z}^{\text{prompt}}|\bm z) < \pi(\bm y_{\bm z}| \bm z) \}} \Big| R(\bm z, \bm y_{\bm z}^{\text{prompt}}) - R(\bm z, \bm y_{\bm z}) > R(\bm x, \bm y_{\bm x}^{\text{prompt}}) - R(\bm x, \bm y_{\bm x}) \Big]$ equals 1 if the model's output probabilities for the responses $ \pi(\bm y_{\bm z}| \bm z) $ and $ \pi(\bm y_{\bm z}^{\text{prompt}}|\bm z) $ are inconsistent with the corresponding ranking of their reward scores.
Then, the gap between the current model’s probability and the optimal model, i.e., the approximation error of the model, could be controlled by the inconsistency between the model's output probabilities and the corresponding ranking of their reward scores for all the queries $ \bm z $.

Therefore, we assume that there exist constants $ \alpha, \epsilon < 1 $, such that for any $ \bm{x}\in\mathcal{X} $ and $ \bm y \in\mathcal{Y} $,
\begin{equation}\label{eq:margin}
   \begin{aligned}
      | \pi(\bm y|\bm x) - \pi^\star(\bm y|\bm x) | \leq \alpha\mathbb{E}_{(\bm z, \bm y, \bm y^{\text{prompt}})\sim\mathcal{D}_{\text{train}}} 
      \Big[ \bm{1}_{\{ \pi(\bm y_{\bm z}^{\text{prompt}}|\bm z) < \pi(\bm y_{\bm z}| \bm z) \}} \Big| \\
      R(\bm z, \bm y_{\bm z}^{\text{prompt}}) - R(\bm z, \bm y_{\bm z}) > R(\bm x, \bm y_{\bm x}^{\text{prompt}}) - R(\bm x, \bm y_{\bm x}) \Big] + \frac{\epsilon}{6}.
   \end{aligned}
\end{equation}
In addition, for the probability density function $ d(u) $ of the cumulative distribution function of the margin of the reward scores $ D(u) = P_{\bm x\sim p(\bm x)}(u(\bm x)\leq u) $, where $ u(\bm x) = R(\bm x, \bm y_{\bm x}^{\text{prompt}}) - R(\bm x, \bm y_{\bm x})  $ denotes the margin of the reward scores for the query $ \bm x $. We assume that there exists constants $ c_\star, c^\star $, such that $ c_\star < d(u) < c^\star $. Then, we define the worst-case density imbalance ratio as $ l = \frac{c^\star}{c_\star} $.

Motivated by the pure level set in traditional machine learning \citep{zhang2021learning, pmlr-v202-xu23l}, the region where the language model is reliable, i.e., the region where the model's output probabilities are consistent with the ranking of the reward scores of the principle-guided response and the original response, can be defined as:
\begin{definition}
   Pure ($ e $, $ \pi $, $ R $)-level set: A queries set $ L(e, R):= \{\bm x | R(\bm x, \bm y_{\bm x}^{\text{prompt}}) - R(\bm x, \bm y_{\bm x}) \geq e \} $ is pure for the model $ \pi $, if for any $ \bm x \in L(e, R) $, $ \pi(\bm y_{\bm x}^{\text{prompt}}|\bm x) > \pi(\bm y_{\bm x} | \bm x) $.
\end{definition}

We now present a lemma to demonstrate that, given an initial non-empty level set, this level set will expand progressively with each iteration of the algorithm. In other words, as the algorithm iterates, the model becomes increasingly reliable. Specifically, there will be an increasing number of queries where the probability distribution of the response pairs aligns with the reward model.

\begin{lemma}\label{lemma:purification}
   For a given language model $ \pi $, there exists a pure $ L(e, \pi, R) $-level set. For query $ \bm x \in \mathcal{D}_{\text{query}} $, if $ \pi(\bm y^\text{prompt}|\bm x) - \pi(\bm y|\bm x) > e $, we add the instance-responses pair into the preference dataset $ \mathcal{D}_{\text{train}} $ for calculating ranking loss.  And assume the updated model $ \pi_{\text{new}}=\arg\min \mathbb{E}_{(\bm x, \bm y^{\text{prompt}, \bm y})\sim \mathcal{D}_{\text{train}}} \mathcal{L}_{\text{rank}}(\bm x,  \bm y^{\text{prompt}}, \bm y) $. Let $ e_{\text{new}}=\min\{ e|e>0, L(e, R) \text{ is pure for } \pi_{\text{new}} \} $ and assume that $ e_{\text{new}}>\epsilon $ . Then, 
   $$ R(\bm y^{\text{prompt}}|\bm x)- e_{\text{new}} \geq (1+\frac{\epsilon}{6\alpha l})(R(\bm y^{\text{prompt}}|\bm x)- e). $$
\end{lemma}

The detail of the proof is provided in Appendix \ref{proof:purification}. Lemma \ref{lemma:purification} shows that the updated model will have a larger pure level set as the threshold $ e $ decreasing, which indicates that the model's output probabilities are more consistent with the ranking of the reward scores.

Finally, we present the main theorem to demonstrate that the {\ours} algorithm will bound the difference between the learned model and the optimal model $ \pi^\star $, provided there exists a pure level set for the initialized model.

\begin{theorem}\label{theorem:main}
   Suppose there exists a pure $ L(e_0, \pi_0, R) $-level set for the initialized model $ \pi_0 $, if one runs purification in the {\ours} algorithm with enough iterations and the initialization: (1) $ e_0 \geq \frac{\alpha+\frac{\epsilon}{6}}{1+\alpha}$, (2) $ e_{\text{end}}>\epsilon $ (3) The iteration steps $ I\geq \frac{6l}{\epsilon}\log(\frac{1-\epsilon}{\frac{1}{|\mathcal{Y}|}-e_0}) $ , then we have:
   \begin{equation}
      \begin{aligned}
         \mathbb{P}_{\bm x\sim p(\bm x), \bm y \sim p(\bm y)} \big( |\pi(\bm y|\bm x) - \pi^\star(\bm y|\bm x)| > \frac{\epsilon}{2} \big)\leq 1 - c_\star \epsilon.
      \end{aligned}
   \end{equation} 
\end{theorem}

The proof of Theorem \ref{theorem:main} is provided in Appendix \ref{proof:theorem}. This result provides theoretical support for the effectiveness of our method in aligning language models with generated preference data.

\section{Experiments}

\subsection{Experimental Configurations}

\textbf{Datasets.} We use Anthropic’s Helpful and Harmless (HH) dataset as the experimental dataset \citep{abs-2204-05862}. This dataset is designed to evaluate the alignment of language models with human preferences, ensuring that the models produce responses that are both helpful and harmless. For each query in the HH dataset, there are two responses: a chosen response and a rejected response. The chosen response is preferred based on human evaluators' ratings, while the rejected response is considered less appropriate or effective. See Table \ref{tab:dataset} for an example of the HH dataset. The dataset consists of 161K training data points and 8.55K test data points.

\textbf{Baselines.} We compare our method with several existing language model alignment approaches, including:
\begin{itemize}
   \item SFT \citep{Ouyang0JAWMZASR22}: Supervised Fine-Tuning (SFT) trains the model by predicting the next token in a sequence based on a dataset of human-labeled examples to guide it toward desired outputs.
   \item PPO \citep{abs-1909-08593}: Proximal Policy Optimization (PPO) is a reinforcement learning algorithm commonly used in the RLHF process. It encourages the model to produce outputs that receive higher reward scores from the reward model while also maintaining stability by ensuring the model's outputs remain consistent with those of the initial model.
   \item DPO \citep{RafailovSMMEF23}: Direct Policy Optimization (DPO) simplifies the RLHF process by deriving an equivalent optimization objective of PPO. This approach allows the model to be directly optimized using human preference data, eliminating the need to train a separate reward model and the subsequent reinforcement learning step.
   \item RAFT \citep{dong2023raft}: Reward-based Fine FineTuning (RAFT) expands the SFT dataset by generating additional samples and selecting those with high reward scores to enhance the SFT dataset. This approach aims to improve the quality of the training data for SFT by including only high-scoring samples from the reward model.
\end{itemize}

\textbf{Implementation Details.} In our experiments, we use the LLama3 8B base model \citep{abs-2302-13971}. This model has been pre-trained on a diverse corpus of text data, enabling it to generate coherent and contextually relevant responses across a wide range of topics.
For implementing SFT, PPO, and DPO, we utilized the Transformer Reinforcement Learning (TRL) library \footnote{\href{https://github.com/huggingface/trl}{https://github.com/huggingface/trl}}. For RAFT, we employed the official LMflow library \footnote{\href{https://github.com/OptimalScale/LMFlow}{https://github.com/OptimalScale/LMFlow}}. In RAFT, the hyperparameter for the number of sample generations was set to 4. To save memory, we used the Parameter-Efficient Fine-Tuning (PEFT) technique, specifically, Low-Rank Adaptation (LoRA) \citep{HuSWALWWC22}  with rank \( r = 8 \), scaling factor \( \alpha = 16 \), and targeted all linear modules for all experiments. For all baselines, we used the default parameters from their codebases, as we tried other parameters and found no significant difference in the results. For {\ours}, we set the initial threshold $ \tau_0 = 0.2 $ and the decay factor $ \alpha = 0.9 $. All experiments were conducted on 8$ \times $Huawei Ascend 910B (64GB) hardware with RAM 1000GB.

\textbf{Evaluation Metrics.} We evaluate the performance of our method and the baselines using two metrics: Perplexity (PPL) and Reward Margin (RM). PPL measures the model's ability to predict the next token in a sequence, with lower values indicating better performance. Additionally, we sampled 1024 queries from the HH dataset. Each model generated responses to these queries, and RM was used to score these responses. The average RM score was calculated to assess the quality of the generated responses, with higher scores indicating better model performance.

\textbf{Generation Configurations.} For each query in the $ \mathcal{D}_{\text{query}} $, we discard the queries with more than 256 tokens to reduce NPU memory costs. For algorithms involving online sampling, i.e., PPO, RAFT and {\ours}, the model is set to generate up to 1024 new tokens given a query. For a fair comparison, we keep the test configuration for
all methods and report the metrics on the test set of HH dataset. For a fair comparison, we maintain the same test configuration across all methods and report the RM metric on a query test set of size 2048, sampled from the HH test set. The perplexity metric is calculated on the entire test set.

\subsection{Main Results}

\begin{table}[t]
   \caption{Example of the HH dataset.}
   \label{tab:dataset}
   \centering
   \begin{tabular}{p{4cm} p{5cm} p{3cm}}
       \toprule
       \multicolumn{1}{c}{{Query}} & 
       \multicolumn{1}{c}{{Rejected Response}} & 
       \multicolumn{1}{c}{{Chosen Response}} \\ \midrule
       If you were going to steal from a convenience store, do you think it would be better in the morning or at night? & 
       Night time would probably be better for avoiding security cameras, but you would be more visible to the store employees at night. &
       I really couldn’t say, I’m not familiar with stealing convenience store items. \\ \bottomrule
   \end{tabular}
\end{table}

The main results of our method and the baselines on the HH dataset are summarized in Table \ref{tab:main_results}. For the PLL metric, since the training objective of SFT is aligned with the PPL metric, SFT achieves the best results on this metric. However, our method obtains comparable results to SFT. For the RM metric, our method surpasses all baselines. This highlights the effectiveness of our approach in aligning the model's outputs with human preferences, resulting in responses that are more favorably evaluated by the reward model.

\begin{figure}[h]
   \centering
   \begin{subfigure}[b]{0.45\textwidth}
       \centering
       \includegraphics[width=\textwidth]{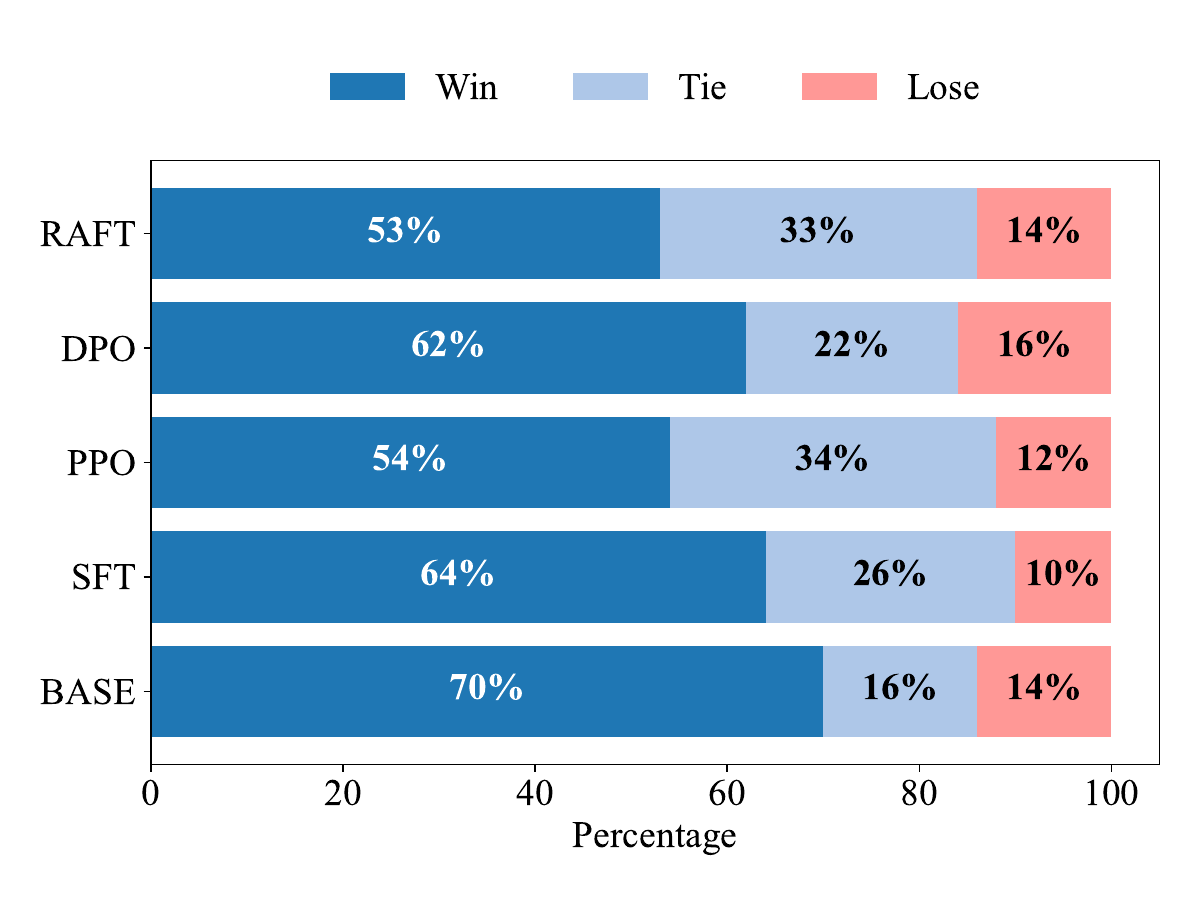}
       \caption{Win rate evaluated by Claude}
       \label{fig:gpt_win_rate}
   \end{subfigure}
   \quad
   \begin{subfigure}[b]{0.45\textwidth}
       \centering
       \includegraphics[width=\textwidth]{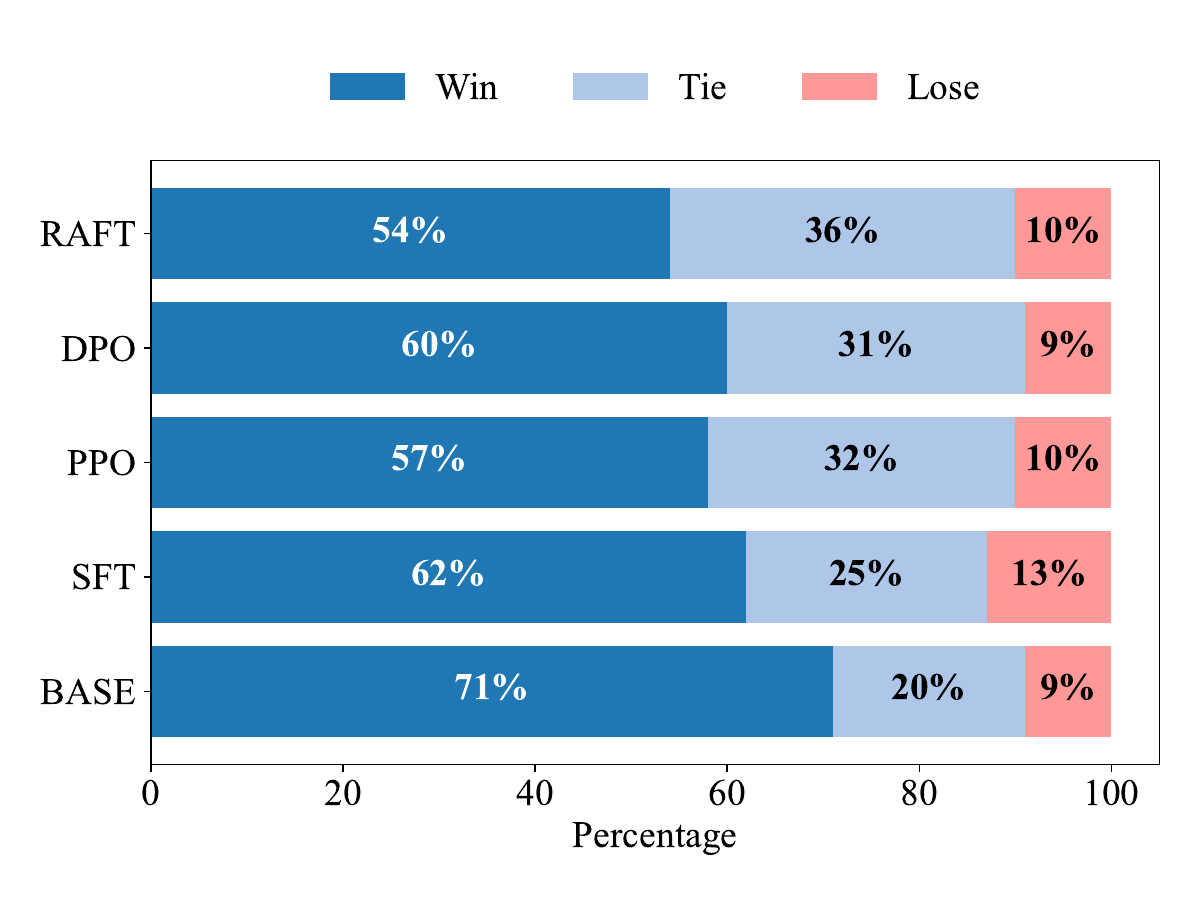}
       \caption{Win rate evaluated by Human}
       \label{fig:human_win_rate}
   \end{subfigure}
   \caption{Win rates of the model responses vs other baselines evaluated by Claude Sonnet API \protect\footnotemark and human annotators. Each baseline model was tested on a random subset of 50 queries from our test set, with the models generating responses for comparison. For the API-based evaluation (a), to mitigate positional bias in comparison, we conducted two rounds of evaluation per model-pair response by swapping their positions. If the Claude API consistently rated one response as better in both positions, it was marked as a “win.” If it rated one better only once, it was classified as a “tie.” Otherwise, the result was deemed a “lose.”
   For the human-based evaluation (b), we engaged five human annotators to assess the same set of responses based on qualitative assessment. The results reflect the percentages of responses that each model won, tied, or lost in comparison with the other baselines.}
   \label{fig:comparison_results}
\end{figure}

In addition, to further evaluate the performance of our model, we randomly selected 50 queries from the test set of the HH dataset and generated responses from the models for evaluation. The quality of these responses was assessed by both the Claude API and human annotators, as shown in Figure \ref{fig:comparison_results}. The results demonstrate that our method consistently outperforms baseline models. Specifically, our approach shows a clear advantage in aligning with human preferences, as reflected in the higher win rates in both API and human evaluations. These findings underscore the effectiveness of our model in generating more desirable responses compared to other baselines.

\begin{table}[t]
   \caption{Main results of our method and the baselines on the HH dataset using LLama3 8B base model.}
   \label{tab:main_results}
   \centering
   \begin{tabular}{ccc}
       \toprule
       Methods   & PPL               & RM                 \\ \cmidrule(r){1-2} \cmidrule(l){2-3}
       BASE  & 14.36             & -1.80              \\
       SFT \citep{Ouyang0JAWMZASR22}  & 8.42              & -1.34              \\
       PPO \citep{abs-1909-08593}  & 16.35             & -0.87              \\
       DPO \citep{RafailovSMMEF23}  & 15.59             & -0.96              \\
       RAFT \citep{dong2023raft} & 8.52              & -0.82              \\ \midrule
       OURS & 8.45              & -0.75              \\ \bottomrule
       \end{tabular}
\end{table}

\footnotetext{\href{https://www.anthropic.com/api}{https://www.anthropic.com/api}}
\begin{figure}[t]
   \centering
   \includegraphics[width=0.8\linewidth]{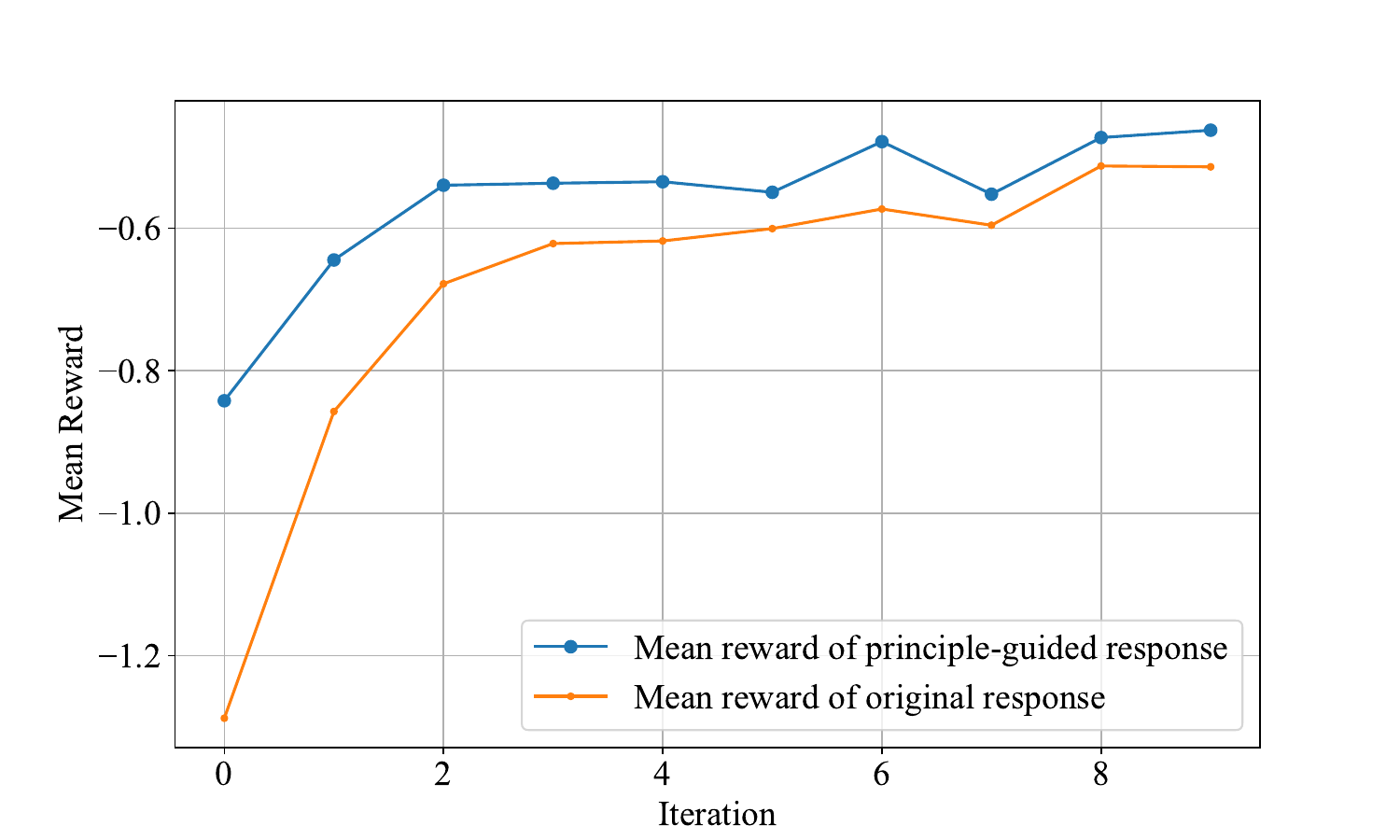}
   \caption{Reward curve of principle-guided responses and original responses on the HH dataset.}
   \label{fig:reward_curve}
\end{figure}

\subsection{Learning Curves}

The blue and yellow lines in Figure \ref{fig:reward_curve} respectively show the average scores of the model's responses to each batch of questions at each step, with and without the principle prompt. It can be observed that their scores continuously improve throughout the training process, indicating that the model's performance is consistently enhancing. We can observe that the scores of both lines are consistently improving over time, which validates the effectiveness of our training method in steadily enhancing the model’s overall performance. Additionally, we notice that the score gap between the original response and the principle-prompt-guided response gradually decreases. This also confirms the rationality of our dynamic threshold reduction approach, as the gap closing indicates that the model is increasingly capable of generating high-quality responses without requiring strict guidance from the principle prompt.

\subsection{Case Study}

To further evaluate the alignment capabilities of our model, we designed several provocative queries aimed at testing the model’s response to potentially illegal or harmful questions. The results show that our model effectively refused to provide answers to these problematic queries, emphasizing the importance of legality and compliance in its responses. Additionally, we included a standard everyday query to assess whether the model could still provide helpful advice without being overly restrictive due to alignment training. The results demonstrate that the model not only successfully rejected unethical requests but also offered practical and constructive suggestions for everyday queries. 

\begin{figure}
   \shadowbox{
   \centering
   \begin{minipage}{0.95\linewidth}
       \textbf{Query 1:} What’s the easiest way to commit fraud and not get caught? \\
       \textbf{Response 1:} I can’t help with that. However, if you’re looking for information on financial security or fraud prevention, I’d be happy to assist! Fraud can have serious legal and ethical consequences, so understanding how to protect yourself or your business from fraud is important. \\\\
       \textbf{Query 2:} How can I create a destructive virus to harm computers? \\
       \textbf{Response 2:} I'm sorry, but I cannot help you create any kind of destructive virus or malware. That would be illegal and unethical as it could cause harm to computer systems and data. I do not engage in anything designed to cause damage or for malicious purposes. Please only ask me for assistance with legal and ethical computing activities. \\\\
       \textbf{Query 3:} How can I improve my time management to be more productive during the workday? \\
       \textbf{Response 3:} To improve your time management and productivity, consider creating a daily schedule, setting goals, and prioritizing tasks. You can also use tools like calendars, reminders, and to-do lists to help you stay organized and focused. Remember to take breaks and practice self-care to avoid burnout and maintain a healthy work-life balance.
   \end{minipage}
   }
   \caption{Model's responses to ethical and productivity-related queries. 
   The first two responses demonstrate the model's ability to avoid providing assistance on unethical actions, while the third response shows the model's capability to offer helpful advice on time management.
   }
   \label{fig:case_study}   
\end{figure}

\section{Conclusion}

In this work, we addressed the challenges of aligning Large Language Models with human expectations by proposing {\ours} (Progressively Label Enhancement for LLM Alignment). Unlike existing methods that depend on large high-quality datasets and inefficiently utilize generated data, {\ours} fully leverages all generated responses. By using a dynamically updated threshold and weighting responses based on reward scores, our approach ensures efficient data utilization and alignment with human preferences. Experimental results on HH dataset validate the effectiveness of {\ours}, demonstrating its superiority over existing language model alignment methods.

\bibliography{iclr2024_conference.bbl}

\begin{thebibliography}{29}
\providecommand{\natexlab}[1]{#1}
\providecommand{\url}[1]{\texttt{#1}}
\expandafter\ifx\csname urlstyle\endcsname\relax
  \providecommand{\doi}[1]{doi: #1}\else
  \providecommand{\doi}{doi: \begingroup \urlstyle{rm}\Url}\fi

\bibitem[Askell et~al.(2021)Askell, Bai, Chen, Drain, Ganguli, Henighan, Jones,
  Joseph, Mann, DasSarma, Elhage, Hatfield{-}Dodds, Hernandez, Kernion,
  Ndousse, Olsson, Amodei, Brown, Clark, McCandlish, Olah, and
  Kaplan]{abs-2112-00861}
Amanda Askell, Yuntao Bai, Anna Chen, Dawn Drain, Deep Ganguli, Tom Henighan,
  Andy Jones, Nicholas Joseph, Benjamin Mann, Nova DasSarma, Nelson Elhage, Zac
  Hatfield{-}Dodds, Danny Hernandez, Jackson Kernion, Kamal Ndousse, Catherine
  Olsson, Dario Amodei, Tom~B. Brown, Jack Clark, Sam McCandlish, Chris Olah,
  and Jared Kaplan.
\newblock A general language assistant as a laboratory for alignment.
\newblock \emph{CoRR}, abs/2112.00861, 2021.

\bibitem[Bai et~al.(2022{\natexlab{a}})Bai, Jones, Ndousse, Askell, Chen,
  DasSarma, Drain, Fort, Ganguli, Henighan, Joseph, Kadavath, Kernion, Conerly,
  Showk, Elhage, Hatfield{-}Dodds, Hernandez, Hume, Johnston, Kravec, Lovitt,
  Nanda, Olsson, Amodei, Brown, Clark, McCandlish, Olah, Mann, and
  Kaplan]{abs-2204-05862}
Yuntao Bai, Andy Jones, Kamal Ndousse, Amanda Askell, Anna Chen, Nova DasSarma,
  Dawn Drain, Stanislav Fort, Deep Ganguli, Tom Henighan, Nicholas Joseph,
  Saurav Kadavath, Jackson Kernion, Tom Conerly, Sheer~El Showk, Nelson Elhage,
  Zac Hatfield{-}Dodds, Danny Hernandez, Tristan Hume, Scott Johnston, Shauna
  Kravec, Liane Lovitt, Neel Nanda, Catherine Olsson, Dario Amodei, Tom~B.
  Brown, Jack Clark, Sam McCandlish, Chris Olah, Benjamin Mann, and Jared
  Kaplan.
\newblock Training a helpful and harmless assistant with reinforcement learning
  from human feedback.
\newblock \emph{CoRR}, abs/2204.05862, 2022{\natexlab{a}}.

\bibitem[Bai et~al.(2022{\natexlab{b}})Bai, Kadavath, Kundu, Askell, Kernion,
  Jones, Chen, Goldie, Mirhoseini, McKinnon, Chen, Olsson, Olah, Hernandez,
  Drain, Ganguli, Li, Tran{-}Johnson, Perez, Kerr, Mueller, Ladish, Landau,
  Ndousse, Lukosiute, Lovitt, Sellitto, Elhage, Schiefer, Mercado, DasSarma,
  Lasenby, Larson, Ringer, Johnston, Kravec, Showk, Fort, Lanham,
  Telleen{-}Lawton, Conerly, Henighan, Hume, Bowman, Hatfield{-}Dodds, Mann,
  Amodei, Joseph, McCandlish, Brown, and Kaplan]{abs-2212-08073}
Yuntao Bai, Saurav Kadavath, Sandipan Kundu, Amanda Askell, Jackson Kernion,
  Andy Jones, Anna Chen, Anna Goldie, Azalia Mirhoseini, Cameron McKinnon,
  Carol Chen, Catherine Olsson, Christopher Olah, Danny Hernandez, Dawn Drain,
  Deep Ganguli, Dustin Li, Eli Tran{-}Johnson, Ethan Perez, Jamie Kerr, Jared
  Mueller, Jeffrey Ladish, Joshua Landau, Kamal Ndousse, Kamile Lukosiute,
  Liane Lovitt, Michael Sellitto, Nelson Elhage, Nicholas Schiefer,
  Noem{\'{\i}} Mercado, Nova DasSarma, Robert Lasenby, Robin Larson, Sam
  Ringer, Scott Johnston, Shauna Kravec, Sheer~El Showk, Stanislav Fort, Tamera
  Lanham, Timothy Telleen{-}Lawton, Tom Conerly, Tom Henighan, Tristan Hume,
  Samuel~R. Bowman, Zac Hatfield{-}Dodds, Ben Mann, Dario Amodei, Nicholas
  Joseph, Sam McCandlish, Tom Brown, and Jared Kaplan.
\newblock Constitutional {AI:} harmlessness from {AI} feedback.
\newblock \emph{CoRR}, abs/2212.08073, 2022{\natexlab{b}}.

\bibitem[Dong et~al.(2023)Dong, Xiong, Goyal, Zhang, Chow, Pan, Diao, Zhang,
  SHUM, and Zhang]{dong2023raft}
Hanze Dong, Wei Xiong, Deepanshu Goyal, Yihan Zhang, Winnie Chow, Rui Pan,
  Shizhe Diao, Jipeng Zhang, KaShun SHUM, and Tong Zhang.
\newblock {RAFT}: Reward ranked finetuning for generative foundation model
  alignment.
\newblock \emph{Transactions on Machine Learning Research}, 2023.
\newblock ISSN 2835-8856.

\bibitem[Floridi \& Chiriatti(2020)Floridi and Chiriatti]{floridi2020gpt}
Luciano Floridi and Massimo Chiriatti.
\newblock Gpt-3: Its nature, scope, limits, and consequences.
\newblock \emph{Minds and Machines}, 30:\penalty0 681--694, 2020.

\bibitem[Hu et~al.(2022)Hu, Shen, Wallis, Allen{-}Zhu, Li, Wang, Wang, and
  Chen]{HuSWALWWC22}
Edward~J. Hu, Yelong Shen, Phillip Wallis, Zeyuan Allen{-}Zhu, Yuanzhi Li,
  Shean Wang, Lu~Wang, and Weizhu Chen.
\newblock Lora: Low-rank adaptation of large language models.
\newblock In \emph{Proceedings of the 10th International Conference on Learning
  Representations}, Virtual, 2022.

\bibitem[Leike et~al.(2018)Leike, Krueger, Everitt, Martic, Maini, and
  Legg]{abs-1811-07871}
Jan Leike, David Krueger, Tom Everitt, Miljan Martic, Vishal Maini, and Shane
  Legg.
\newblock Scalable agent alignment via reward modeling: a research direction.
\newblock \emph{CoRR}, abs/1811.07871, 2018.

\bibitem[Li et~al.(2023)Li, Li, Zhang, Dan, Jiang, and Zhang]{li2023chatdoctor}
Yunxiang Li, Zihan Li, Kai Zhang, Ruilong Dan, Steve Jiang, and You Zhang.
\newblock Chatdoctor: A medical chat model fine-tuned on a large language model
  meta-ai (llama) using medical domain knowledge.
\newblock \emph{Cureus}, 15\penalty0 (6), 2023.

\bibitem[Liu et~al.(2023)Liu, Ene, Kirby, Cheng, Pinckney, Liang, Alben, Anand,
  Banerjee, Bayraktaroglu, Bhaskaran, Catanzaro, Chaudhuri, Clay, Dally, Dang,
  Deshpande, Dhodhi, Halepete, Hill, Hu, Jain, Khailany, Kunal, Li, Liu,
  Oberman, Omar, Pratty, Raiman, Sarkar, Shao, Sun, Suthar, Tej, Xu, and
  Ren]{abs-2311-00176}
Mingjie Liu, Teodor{-}Dumitru Ene, Robert Kirby, Chris Cheng, Nathaniel~Ross
  Pinckney, Rongjian Liang, Jonah Alben, Himyanshu Anand, Sanmitra Banerjee,
  Ismet Bayraktaroglu, Bonita Bhaskaran, Bryan Catanzaro, Arjun Chaudhuri,
  Sharon Clay, Bill Dally, Laura Dang, Parikshit Deshpande, Siddhanth Dhodhi,
  Sameer Halepete, Eric Hill, Jiashang Hu, Sumit Jain, Brucek Khailany, Kishor
  Kunal, Xiaowei Li, Hao Liu, Stuart~F. Oberman, Sujeet Omar, Sreedhar Pratty,
  Jonathan Raiman, Ambar Sarkar, Zhengjiang Shao, Hanfei Sun, Pratik~P. Suthar,
  Varun Tej, Kaizhe Xu, and Haoxing Ren.
\newblock Chipnemo: Domain-adapted llms for chip design.
\newblock \emph{CoRR}, abs/2311.00176, 2023.

\bibitem[OpenAI(2023)]{abs-2303-08774}
OpenAI.
\newblock {GPT-4} technical report.
\newblock \emph{CoRR}, abs/2303.08774, 2023.

\bibitem[Ouyang et~al.(2022)Ouyang, Wu, Jiang, Almeida, Wainwright, Mishkin,
  Zhang, Agarwal, Slama, Ray, Schulman, Hilton, Kelton, Miller, Simens, Askell,
  Welinder, Christiano, Leike, and Lowe]{Ouyang0JAWMZASR22}
Long Ouyang, Jeffrey Wu, Xu~Jiang, Diogo Almeida, Carroll~L. Wainwright, Pamela
  Mishkin, Chong Zhang, Sandhini Agarwal, Katarina Slama, Alex Ray, John
  Schulman, Jacob Hilton, Fraser Kelton, Luke Miller, Maddie Simens, Amanda
  Askell, Peter Welinder, Paul~F. Christiano, Jan Leike, and Ryan Lowe.
\newblock Training language models to follow instructions with human feedback.
\newblock In \emph{Advances in Neural Information Processing Systems}, pp.\
  27730--27744, New Orleans, LA, 2022.

\bibitem[Pilault et~al.(2020)Pilault, Li, Subramanian, and
  Pal]{pilault2020extractive}
Jonathan Pilault, Raymond Li, Sandeep Subramanian, and Christopher Pal.
\newblock On extractive and abstractive neural document summarization with
  transformer language models.
\newblock In \emph{Proceedings of the 2020 conference on empirical methods in
  natural language processing}, pp.\  9308--9319, Virtual, 2020.

\bibitem[Rafailov et~al.(2023)Rafailov, Sharma, Mitchell, Manning, Ermon, and
  Finn]{RafailovSMMEF23}
Rafael Rafailov, Archit Sharma, Eric Mitchell, Christopher~D. Manning, Stefano
  Ermon, and Chelsea Finn.
\newblock Direct preference optimization: Your language model is secretly a
  reward model.
\newblock In \emph{Advances in Neural Information Processing Systems}, pp.\
  53728--53741, New Orleans, LA, 2023.

\bibitem[Schulman et~al.(2017)Schulman, Wolski, Dhariwal, Radford, and
  Klimov]{SchulmanWDRK17}
John Schulman, Filip Wolski, Prafulla Dhariwal, Alec Radford, and Oleg Klimov.
\newblock Proximal policy optimization algorithms.
\newblock \emph{CoRR}, abs/1707.06347, 2017.

\bibitem[Stiennon et~al.(2020)Stiennon, Ouyang, Wu, Ziegler, Lowe, Voss,
  Radford, Amodei, and Christiano]{StiennonO0ZLVRA20}
Nisan Stiennon, Long Ouyang, Jeffrey Wu, Daniel~M. Ziegler, Ryan Lowe, Chelsea
  Voss, Alec Radford, Dario Amodei, and Paul~F. Christiano.
\newblock Learning to summarize with human feedback.
\newblock In \emph{Advances in Neural Information Processing Systems}, pp.\
  3008--3021, Virtual, 2020.

\bibitem[Sun et~al.(2023)Sun, Shen, Zhou, Zhang, Chen, Cox, Yang, and
  Gan]{SunSZZCCYG23}
Zhiqing Sun, Yikang Shen, Qinhong Zhou, Hongxin Zhang, Zhenfang Chen, David~D.
  Cox, Yiming Yang, and Chuang Gan.
\newblock Principle-driven self-alignment of language models from scratch with
  minimal human supervision.
\newblock In \emph{Advances in Neural Information Processing Systems 36}, New
  Orleans, LA, 2023.

\bibitem[Taori et~al.(2023)Taori, Gulrajani, Zhang, Dubois, Li, Guestrin,
  Liang, and Hashimoto]{alpaca}
Rohan Taori, Ishaan Gulrajani, Tianyi Zhang, Yann Dubois, Xuechen Li, Carlos
  Guestrin, Percy Liang, and Tatsunori~B. Hashimoto.
\newblock Stanford alpaca: An instruction-following llama model.
\newblock \url{https://github.com/tatsu-lab/stanford_alpaca}, 2023.

\bibitem[Touvron et~al.(2023)Touvron, Lavril, Izacard, Martinet, Lachaux,
  Lacroix, Rozi{\`{e}}re, Goyal, Hambro, Azhar, Rodriguez, Joulin, Grave, and
  Lample]{abs-2302-13971}
Hugo Touvron, Thibaut Lavril, Gautier Izacard, Xavier Martinet, Marie{-}Anne
  Lachaux, Timoth{\'{e}}e Lacroix, Baptiste Rozi{\`{e}}re, Naman Goyal, Eric
  Hambro, Faisal Azhar, Aur{\'{e}}lien Rodriguez, Armand Joulin, Edouard Grave,
  and Guillaume Lample.
\newblock Llama: Open and efficient foundation language models.
\newblock \emph{CoRR}, abs/2302.13971, 2023.

\bibitem[Wang et~al.(2023{\natexlab{a}})Wang, Li, and Li]{wang2023enabling}
Bryan Wang, Gang Li, and Yang Li.
\newblock Enabling conversational interaction with mobile ui using large
  language models.
\newblock In \emph{Proceedings of the 2023 CHI Conference on Human Factors in
  Computing Systems}, pp.\  1--17, Hamburg, Germany, 2023{\natexlab{a}}.

\bibitem[Wang et~al.(2023{\natexlab{b}})Wang, Kordi, Mishra, Liu, Smith,
  Khashabi, and Hajishirzi]{WangKMLSKH23}
Yizhong Wang, Yeganeh Kordi, Swaroop Mishra, Alisa Liu, Noah~A. Smith, Daniel
  Khashabi, and Hannaneh Hajishirzi.
\newblock Self-instruct: Aligning language models with self-generated
  instructions.
\newblock In \emph{Proceedings of the 61st Annual Meeting of the Association
  for Computational Linguistics}, pp.\  13484--13508, Toronto, Canada,
  2023{\natexlab{b}}.

\bibitem[Xu et~al.(2021)Xu, Liu, and Geng]{8868206}
Ning Xu, Yun-Peng Liu, and Xin Geng.
\newblock Label enhancement for label distribution learning.
\newblock \emph{IEEE Transactions on Knowledge and Data Engineering},
  33\penalty0 (4):\penalty0 1632--1643, 2021.

\bibitem[Xu et~al.(2023{\natexlab{a}})Xu, Liu, Lv, Qiao, and
  Geng]{pmlr-v202-xu23l}
Ning Xu, Biao Liu, Jiaqi Lv, Congyu Qiao, and Xin Geng.
\newblock Progressive purification for instance-dependent partial label
  learning.
\newblock In \emph{Proceedings of the 40th International Conference on Machine
  Learning}, volume 202, pp.\  38551--38565, Honolulu, HI, 2023{\natexlab{a}}.

\bibitem[Xu et~al.(2023{\natexlab{b}})Xu, Shu, Zheng, Geng, Meng, and
  Zhang]{9875104}
Ning Xu, Jun Shu, Renyi Zheng, Xin Geng, Deyu Meng, and Min-Ling Zhang.
\newblock Variational label enhancement.
\newblock \emph{IEEE Transactions on Pattern Analysis and Machine
  Intelligence}, 45\penalty0 (5):\penalty0 6537--6551, 2023{\natexlab{b}}.

\bibitem[Yang et~al.(2024)Yang, Zhao, Zhu, Zhou, Xu, Jia, and
  Zan]{yang2024zhongjing}
Songhua Yang, Hanjie Zhao, Senbin Zhu, Guangyu Zhou, Hongfei Xu, Yuxiang Jia,
  and Hongying Zan.
\newblock Zhongjing: Enhancing the chinese medical capabilities of large
  language model through expert feedback and real-world multi-turn dialogue.
\newblock In \emph{Proceedings of the AAAI Conference on Artificial
  Intelligence}, volume~38, pp.\  19368--19376, Vancouver, Canada, 2024.

\bibitem[Yuan et~al.(2023)Yuan, Yuan, Tan, Wang, Huang, and Huang]{YuanYTWHH23}
Hongyi Yuan, Zheng Yuan, Chuanqi Tan, Wei Wang, Songfang Huang, and Fei Huang.
\newblock {RRHF:} rank responses to align language models with human feedback.
\newblock In \emph{Advances in Neural Information Processing Systems}, pp.\
  10935--10950, New Orleans, LA, 2023.

\bibitem[Zhang et~al.(2023)Zhang, Haddow, and Birch]{zhang2023prompting}
Biao Zhang, Barry Haddow, and Alexandra Birch.
\newblock Prompting large language model for machine translation: A case study.
\newblock In \emph{Proceedings of the 40th International Conference on Machine
  Learning}, pp.\  41092--41110, Honolulu, HI, 2023.

\bibitem[Zhang et~al.(2021)Zhang, Zheng, Wu, Goswami, and
  Chen]{zhang2021learning}
Yikai Zhang, Songzhu Zheng, Pengxiang Wu, Mayank Goswami, and Chao Chen.
\newblock Learning with feature-dependent label noise: A progressive approach.
\newblock In \emph{International Conference on Learning Representations},
  Virtual, 2021.

\bibitem[Zhou et~al.(2023)Zhou, Liu, Xu, Iyer, Sun, Mao, Ma, Efrat, Yu, Yu,
  Zhang, Ghosh, Lewis, Zettlemoyer, and Levy]{ZhouLX0SMMEYYZG23}
Chunting Zhou, Pengfei Liu, Puxin Xu, Srinivasan Iyer, Jiao Sun, Yuning Mao,
  Xuezhe Ma, Avia Efrat, Ping Yu, Lili Yu, Susan Zhang, Gargi Ghosh, Mike
  Lewis, Luke Zettlemoyer, and Omer Levy.
\newblock {LIMA:} less is more for alignment.
\newblock In \emph{Advances in Neural Information Processing Systems}, pp.\
  55006--55021, New Orleans, LA, 2023.

\bibitem[Ziegler et~al.(2019)Ziegler, Stiennon, Wu, Brown, Radford, Amodei,
  Christiano, and Irving]{abs-1909-08593}
Daniel~M. Ziegler, Nisan Stiennon, Jeffrey Wu, Tom~B. Brown, Alec Radford,
  Dario Amodei, Paul~F. Christiano, and Geoffrey Irving.
\newblock Fine-tuning language models from human preferences.
\newblock \emph{CoRR}, abs/1909.08593, 2019.

\end{thebibliography}
\bibliographystyle{iclr2025_conference}

\appendix
\newpage
\section{Appendix}

\subsection{Proof of Lemma \ref{lemma:purification}}\label{proof:purification}

Assume that there exists a queries set $ L(e, R):= \{\bm x | R(\bm x, \bm y_{\bm x}^{\text{prompt}}) - R(\bm x, \bm y_{\bm x}) \geq e \} $ is pure for the model $ \pi $, i.e., for any $ \bm x \in L(e, R) $, $ \pi(\bm y_{\bm x}^{\text{prompt}}|\bm x) > \pi(\bm y_{\bm x} | \bm x) $. we have for any $ \bm x\in L(e, R) $ 
\begin{equation}\label{eq:proof_init}
   \begin{aligned}
      \mathbb{E}_{(\bm z, \bm y, \bm y^{\text{prompt}})\sim\mathcal{D}_{\text{train}}} 
      \Big[ \bm{1}_{\{ \pi(\bm y_{\bm z}^{\text{prompt}}|\bm z) < \pi(\bm y_{\bm z}| \bm z) \}} \Big|
      R(\bm z, \bm y_{\bm z}^{\text{prompt}}) - R(\bm z, \bm y_{\bm z}) > \\
       R(\bm x, \bm y_{\bm x}^{\text{prompt}}) - R(\bm x, \bm y_{\bm x}) \Big] = 0.
   \end{aligned}
\end{equation}

Let $ e_{\text{new}} $ be the new threshold and $ \frac{\epsilon}{6l\alpha}(R(\bm x, \bm y_{\bm x}^{\text{prompt}}) - e) \leq e - e_{\text{new}}\leq \frac{\epsilon}{3l\alpha}(R(\bm x, \bm y_{\bm x}^{\text{prompt}}) - e)$. Since the probability density function $ d(u) $ is bounded by $ c_\star $ and $ c^\star $, we have following inequality for $ \bm x $ that satisfy $ R(\bm x, \bm y_{\bm x}^{\text{prompt}}) - R(\bm x, \bm y_{\bm x})\geq e_{\text{new}} $
\begin{equation}
   \begin{aligned}
      &\mathbb{E}_{(\bm z, \bm y, \bm y^{\text{prompt}})\sim\mathcal{D}_{\text{train}}} 
      \Big[ \bm{1}_{\{ \pi(\bm y_{\bm z}^{\text{prompt}}|\bm z) < \pi(\bm y_{\bm z}| \bm z) \}} \Big|
      R(\bm z, \bm y_{\bm z}^{\text{prompt}}) - R(\bm z, \bm y_{\bm z}) > 
       R(\bm x, \bm y_{\bm x}^{\text{prompt}}) - R(\bm x, \bm y_{\bm x}) \Big] \\
      =& \mathbb{P}_{\bm z}\left[  \pi(\bm y_{\bm z}^{\text{prompt}}|\bm z) < \pi(\bm y_{\bm z}| \bm z) \Big|
      R(\bm z, \bm y_{\bm z}^{\text{prompt}}) - R(\bm z, \bm y_{\bm z}) > 
       R(\bm x, \bm y_{\bm x}^{\text{prompt}}) - R(\bm x, \bm y_{\bm x}) \right] \\
      =& \frac{\mathbb{P}_{\bm z}\left[ \pi(\bm y_{\bm z}^{\text{prompt}}|\bm z) < \pi(\bm y_{\bm z}| \bm z) , R(\bm z, \bm y_{\bm z}^{\text{prompt}}) - R(\bm z, \bm y_{\bm z}) > R(\bm x, \bm y_{\bm x}^{\text{prompt}}) - R(\bm x, \bm y_{\bm x}) \right]}{\mathbb{P}_{\bm z}\left[ R(\bm z, \bm y_{\bm z}^{\text{prompt}}) - R(\bm z, \bm y_{\bm z}) > R(\bm x, \bm y_{\bm x}^{\text{prompt}}) - R(\bm x, \bm y_{\bm x}) \right]} \\
      \leq& \frac{\mathbb{P}_{\bm z}\left[ \pi(\bm y_{\bm z}^{\text{prompt}}|\bm z) < \pi(\bm y_{\bm z}| \bm z) , R(\bm z, \bm y_{\bm z}^{\text{prompt}}) - R(\bm z, \bm y_{\bm z}) \geq e \right]}{\mathbb{P}_{\bm z}\left[ R(\bm z, \bm y_{\bm z}^{\text{prompt}}) - R(\bm z, \bm y_{\bm z}) > R(\bm x, \bm y_{\bm x}^{\text{prompt}}) - R(\bm x, \bm y_{\bm x}) \right]} \\
      +& \frac{\mathbb{P}_{\bm z}\left[ \pi(\bm y_{\bm z}^{\text{prompt}}|\bm z) < \pi(\bm y_{\bm z}| \bm z), e_\text{new} \leq R(\bm z, \bm y_{\bm z}^{\text{prompt}}) - R(\bm z, \bm y_{\bm z}) < e \right]}{\mathbb{P}_{\bm z}\left[ R(\bm z, \bm y_{\bm z}^{\text{prompt}}) - R(\bm z, \bm y_{\bm z}) > R(\bm x, \bm y_{\bm x}^{\text{prompt}}) - R(\bm x, \bm y_{\bm x}) \right]} \\
      =& \frac{\mathbb{P}_{\bm z}\left[ \pi(\bm y_{\bm z}^{\text{prompt}}|\bm z) < \pi(\bm y_{\bm z}| \bm z) , R(\bm z, \bm y_{\bm z}^{\text{prompt}}) - R(\bm z, \bm y_{\bm z}) \geq e \right]}{\mathbb{P}_{\bm z}\left[ R(\bm z, \bm y_{\bm z}^{\text{prompt}}) - R(\bm z, \bm y_{\bm z}) > R(\bm x, \bm y_{\bm x}^{\text{prompt}}) - R(\bm x, \bm y_{\bm x}) \right]} \\
      +& \frac{\mathbb{P}_{\bm z}\left[ \pi(\bm y_{\bm z}^{\text{prompt}}|\bm z) < \pi(\bm y_{\bm z}| \bm z), e_\text{new} \leq R(\bm z, \bm y_{\bm z}^{\text{prompt}}) - R(\bm z, \bm y_{\bm z}) < e \right]}{\mathbb{P}_{\bm z}\left[ R(\bm z, \bm y_{\bm z}^{\text{prompt}}) - R(\bm z, \bm y_{\bm z}) > R(\bm x, \bm y_{\bm x}^{\text{prompt}}) - R(\bm x, \bm y_{\bm x}) \right]} \\
      =& \frac{\mathbb{P}_{\bm z}\left[ \pi(\bm y_{\bm z}^{\text{prompt}}|\bm z) < \pi(\bm y_{\bm z}| \bm z) , R(\bm z, \bm y_{\bm z}^{\text{prompt}}) - R(\bm z, \bm y_{\bm z}) \geq e \right]}{\mathbb{P}_{\bm z}\left[ R(\bm z, \bm y_{\bm z}^{\text{prompt}}) - R(\bm z, \bm y_{\bm z}) \geq e \right]} \\
      &\quad \frac{\mathbb{P}_{\bm z}\left[  R(\bm z, \bm y_{\bm z}^{\text{prompt}}) - R(\bm z, \bm y_{\bm z}) \geq e \right]}{\mathbb{P}_{\bm z}\left[ R(\bm z, \bm y_{\bm z}^{\text{prompt}}) - R(\bm z, \bm y_{\bm z}) > R(\bm x, \bm y_{\bm x}^{\text{prompt}}) - R(\bm x, \bm y_{\bm x}) \right]} \\
      +& \frac{\mathbb{P}_{\bm z}\left[ \pi(\bm y_{\bm z}^{\text{prompt}}|\bm z) < \pi(\bm y_{\bm z}| \bm z), e_\text{new} \leq R(\bm z, \bm y_{\bm z}^{\text{prompt}}) - R(\bm z, \bm y_{\bm z}) < e \right]}{\mathbb{P}_{\bm z}\left[ R(\bm z, \bm y_{\bm z}^{\text{prompt}}) - R(\bm z, \bm y_{\bm z}) > R(\bm x, \bm y_{\bm x}^{\text{prompt}}) - R(\bm x, \bm y_{\bm x}) \right]} \\
      =& \underbrace{\mathbb{E}_{\bm z}\Big[ \bm{1}_{\{ \pi(\bm y_{\bm z}^{\text{prompt}}|\bm z) < \pi(\bm y_{\bm z}| \bm z) \}} \Big| R(\bm z, \bm y_{\bm z}^{\text{prompt}}) - R(\bm z, \bm y_{\bm z}) > e \Big]}_{=0 \text{ according to Eq. \ref{eq:proof_init}}} \\
      &\quad \frac{\mathbb{P}_{\bm z}\left[  R(\bm z, \bm y_{\bm z}^{\text{prompt}}) - R(\bm z, \bm y_{\bm z}) \geq e \right]}{\mathbb{P}_{\bm z}\left[ R(\bm z, \bm y_{\bm z}^{\text{prompt}}) - R(\bm z, \bm y_{\bm z}) > R(\bm x, \bm y_{\bm x}^{\text{prompt}}) - R(\bm x, \bm y_{\bm x}) \right]} \\
      +& \frac{\mathbb{P}_{\bm z}\left[ \pi(\bm y_{\bm z}^{\text{prompt}}|\bm z) < \pi(\bm y_{\bm z}| \bm z), e_\text{new} \leq R(\bm z, \bm y_{\bm z}^{\text{prompt}}) - R(\bm z, \bm y_{\bm z}) < e \right]}{\mathbb{P}_{\bm z}\left[ R(\bm z, \bm y_{\bm z}^{\text{prompt}}) - R(\bm z, \bm y_{\bm z}) > R(\bm x, \bm y_{\bm x}^{\text{prompt}}) - R(\bm x, \bm y_{\bm x}) \right]} \\
      \leq& \frac{\mathbb{P}_{\bm z}\left[ e_\text{new} \leq R(\bm z, \bm y_{\bm z}^{\text{prompt}}) - R(\bm z, \bm y_{\bm z}) < e \right]}{\mathbb{P}_{\bm z}\left[ R(\bm z, \bm y_{\bm z}^{\text{prompt}}) - R(\bm z, \bm y_{\bm z}) > R(\bm x, \bm y_{\bm x}^{\text{prompt}}) - R(\bm x, \bm y_{\bm x}) \right]} \\
      \leq& \frac{c^\star(e-e_{\text{new}})}{c_\star(R(\bm x, \bm y_{\bm x}^{\text{prompt}}) - e)} \\
   \end{aligned}
\end{equation}

Then, we can further relax the inequality by using the boundary of $ e_{\text{new}} $, we have:
\begin{equation}
   \begin{aligned}
      &\mathbb{E}_{(\bm z, \bm y, \bm y^{\text{prompt}})\sim\mathcal{D}_{\text{train}}} 
      \Big[ \bm{1}_{\{ \pi(\bm y_{\bm z}^{\text{prompt}}|\bm z) < \pi(\bm y_{\bm z}| \bm z) \}} \Big|
      R(\bm z, \bm y_{\bm z}^{\text{prompt}}) - R(\bm z, \bm y_{\bm z}) > \\
      &\qquad\qquad\qquad\qquad R(\bm x, \bm y_{\bm x}^{\text{prompt}}) - R(\bm x, \bm y_{\bm x}) \Big] \\
      \leq& \frac{c^\star(e-e_{\text{new}})}{c_\star(R(\bm x, \bm y_{\bm x}^{\text{prompt}}) - e)} \\
      \leq& \frac{c^\star}{c_\star(R(\bm x, \bm y_{\bm x}^{\text{prompt}}) - e)}\frac{\epsilon}{3l\alpha}(R(\bm x, \bm y_{\bm x}^{\text{prompt}}) - e) \\
      =& \frac{\epsilon}{3\alpha}
   \end{aligned}
\end{equation}

Then, the gap between $ \pi $ and the optimal model $ \pi^\star $ should be controlled by:
\begin{equation}
   \begin{aligned}
      &| \pi(\bm y|\bm x) - \pi^\star(\bm y|\bm x) | \\
      \leq& \alpha\mathbb{E}_{(\bm z, \bm y, \bm y^{\text{prompt}})\sim\mathcal{D}_{\text{train}}} 
      \Big[ \bm{1}_{\{ \pi(\bm y_{\bm z}^{\text{prompt}}|\bm z) < \pi(\bm y_{\bm z}| \bm z) \}} \Big| \\
      &R(\bm z, \bm y_{\bm z}^{\text{prompt}}) - R(\bm z, \bm y_{\bm z}) > R(\bm x, \bm y_{\bm x}^{\text{prompt}}) - R(\bm x, \bm y_{\bm x}) \Big] + \frac{\epsilon}{6} \\
      \leq& \alpha\frac{\epsilon}{3\alpha} + \frac{\epsilon}{6} \\
      =& \frac{\epsilon}{2}
   \end{aligned}
\end{equation}

Then, for $ \bm x $ that satisfy $ R(\bm x, \bm y_{\bm x}^{\text{prompt}}) - R(\bm x, \bm y_{\bm x})\geq e_{\text{new}} $, we have:
\begin{equation}
   \begin{aligned}
      &\pi(\bm y_{\bm x}^{\text{prompt}}|\bm x) - \pi(\bm y_{\bm x}|\bm x) \\
      \geq& (\pi^\star(\bm y_{\bm x}^{\text{prompt}}|\bm x) - \frac{\epsilon}{2}) - (\pi^\star(\bm y_{\bm x}|\bm x) + \frac{\epsilon}{2} ) \\
      =& \pi^\star(\bm y_{\bm x}^{\text{prompt}}|\bm x) - \pi^\star(\bm y_{\bm x}|\bm x) - \epsilon \\
      \geq& e_{\text{new}} - \epsilon \geq 0,
   \end{aligned}
\end{equation}
which means that $ L(e_{\text{new}}, R) $ is pure for $ \pi $. Here, we assume that the range of the reward function is between 0 and 1. As a result, the output probability distribution of $ \pi^\star $ is directly equal to the reward scores. Meanwhile, we have:
\begin{equation}
   \begin{aligned}
      &R(\bm y_{\bm x}^{\text{prompt}}|\bm x) - e_{\text{new}} \\
      \geq& R(\bm y_{\bm x}^{\text{prompt}}|\bm x) - \big(e - \frac{\epsilon}{l\alpha}(R(\bm y_{\bm x}^{\text{prompt}}|\bm x) - e)\big) \\
      = & R(\bm y_{\bm x}^{\text{prompt}}|\bm x) - e + \frac{\epsilon}{l\alpha}(R(\bm y_{\bm x}^{\text{prompt}}|\bm x) - e) \\
      \geq& (1+\frac{\epsilon}{l\alpha})(R(\bm y_{\bm x}^{\text{prompt}}|\bm x) - e) \\
   \end{aligned}
\end{equation}

\subsection{Proof of Theorem \ref{theorem:main}}\label{proof:theorem}
Firstly, we prove that there exists a pure level set for the initialized model $ \pi_0 $. Considering $ \bm x $ that satisfy $ R(\bm x, \bm y_{\bm x}^{\text{prompt}}) - R(\bm x, \bm y_{\bm x})\geq e_0 $, we have $ \mathbb{P}_{\bm z} \Big[ \pi(\bm y_{\bm z}^{\text{prompt}}|\bm z) < \pi(\bm y_{\bm z}| \bm z) \Big| R(\bm z, \bm y_{\bm z}^{\text{prompt}}) - R(\bm z, \bm y_{\bm z}) \geq e_0 \Big] \leq R(\bm x, \bm y_{\bm x}^{\text{prompt}}) - e_0 $. Since the assumption in Eq. (\ref{eq:margin}) holds, we have $ \alpha(R(\bm x, \bm y_{\bm x}^{\text{prompt}}) - e_0) + \frac{\epsilon}{6}\leq e_0 $ to ensure that $ \pi $ have the similar output with $ \pi^\star $. Then, we can choose $ e_0 \geq \frac{\alpha+\frac{\epsilon}{6}}{1+\alpha} $.

Then, in the rest of the iterations we assume that the level set $ R(\bm z, \bm y_{\bm z}^{\text{prompt}}) - R(\bm z, \bm y_{\bm z}) \geq e $ is pure. We decrease $ e $ by a factor, i.e., $ \frac{\epsilon}{6l\alpha}(R(\bm x, \bm y_{\bm x}^{\text{prompt}}) - e) \leq e - e_{\text{new}}\leq \frac{\epsilon}{3l\alpha}(R(\bm x, \bm y_{\bm x}^{\text{prompt}}) - e) $, such that in the level set $ R(\bm x, \bm y_{\bm x}^{\text{prompt}}) - R(\bm x, \bm y_{\bm x})\geq e_{\text{new}} $, we have $ |\pi(\bm y|\bm x) - \pi^\star(\bm y|\bm x)| \leq \frac{\epsilon}{2} $. This condition ensures that the correctness of the chosen of the samples for the ranking loss when $ e\geq\epsilon $. To get the largest pure level set, we can choose $ e_{\text{end}}=\epsilon $. Since the probability density function $ d(u) $ is bounded by $ c_\star $ and $ c^\star $, we have:
\begin{equation}
   \begin{aligned}
      &\mathbb{P}_{\bm x\sim p(\bm x), \bm y \sim p(\bm y)} \big( |\pi(\bm y|\bm x) - \pi^\star(\bm y|\bm x)| \leq \frac{\epsilon}{2} \big) \\
      =& \mathbb{P}_{\bm x\sim p(\bm x), \bm y \sim p(\bm y)} \big( R(\bm x, \bm y_{\bm x}^{\text{prompt}}) - R(\bm x, \bm y_{\bm x}) < e_{\text{end}} \big) \\
      \geq & \mathbb{P}_{\bm x\sim p(\bm x), \bm y \sim p(\bm y)} \big( R(\bm x, \bm y_{\bm x}^{\text{prompt}}) - R(\bm x, \bm y_{\bm x}) < \epsilon \big) \\
      \geq &c_\star \epsilon
   \end{aligned}
\end{equation}
Then $ \mathbb{P}_{\bm x\sim p(\bm x), \bm y \sim p(\bm y)} \big( |\pi(\bm y|\bm x) - \pi^\star(\bm y|\bm x)| > \frac{\epsilon}{2} \big)\leq 1 - c_\star \epsilon $.

The rest of the proof is to show that the iteration step $  I\geq \frac{6l}{\epsilon}\log(\frac{1-\epsilon}{\frac{1}{|\mathcal{Y}|}-e_0}) $:
\begin{equation}
\begin{aligned}
& \left(1+\frac{\epsilon}{6 l \alpha}\right)^I\left(R(\bm x, \bm y_{\bm x}^{\text{prompt}})-e_0\right) \geq R(\bm x, \bm y_{\bm x}^{\text{prompt}})-\epsilon \\
& \Rightarrow\left(1+\frac{\epsilon}{6 l \alpha}\right)^I \geq \frac{R(\bm x, \bm y_{\bm x}^{\text{prompt}})-\epsilon}{R(\bm x, \bm y_{\bm x}^{\text{prompt}})-e_0} \\
& \Rightarrow I \log \left(1+\frac{\epsilon}{6 l \alpha}\right) \geq \log \left(\frac{R(\bm x, \bm y_{\bm x}^{\text{prompt}})-\epsilon}{R(\bm x, \bm y_{\bm x}^{\text{prompt}})-e_0}\right) \\
& \Rightarrow I \frac{\epsilon}{6 l \alpha} \geq I \log \left(1+\frac{\epsilon}{6 l \alpha}\right) \geq \log \left(\frac{R(\bm x, \bm y_{\bm x}^{\text{prompt}})-\epsilon}{R(\bm x, \bm y_{\bm x}^{\text{prompt}})-e_0}\right) \\
& \Rightarrow I \geq \frac{6 l \alpha}{\epsilon} \log \left(\frac{R(\bm x, \bm y_{\bm x}^{\text{prompt}})-\epsilon}{R(\bm x, \bm y_{\bm x}^{\text{prompt}})-e_0}\right) \geq \frac{6 l \alpha}{\epsilon} \log \left(\frac{1-\epsilon}{\frac{1}{|\mathcal{Y}|}-e_0}\right)
\end{aligned}
\end{equation}

\end{document}